\documentclass[conference]{IEEEtran}
\IEEEoverridecommandlockouts
\usepackage{cite}
\usepackage{amsmath,amssymb,amsfonts}
\usepackage{algorithmic}
\usepackage{booktabs}
\usepackage{graphicx}
\usepackage{array}  
\usepackage{tabularx}
\usepackage{caption}
\usepackage{afterpage}
\usepackage{tikz}
\usepackage{adjustbox}
\usepackage{textcomp}
\usepackage{xcolor}
\usepackage{float}
\usepackage{placeins}
\usetikzlibrary{positioning, shapes, arrows}
\usepackage[numbers,sort&compress]{natbib}
\usepackage[colorlinks=true, linkcolor=blue, citecolor=blue, urlcolor=blue]{hyperref}
\def\BibTeX{{\rm B\kern-.05em{\sc i\kern-.025em b}\kern-.08em
    T\kern-.1667em\lower.7ex\hbox{E}\kern-.125emX}}
\begin{document}

\title{Based on Data Balancing and Model Improvement for Multi-Label Sentiment Classification Performance Enhancement study}
\author{
\footnotesize
\begin{tabular}{c c c c}
1\textsuperscript{st} Zijin Su* & 1\textsuperscript{st} Huanzhu Lv & 2\textsuperscript{nd} Yuren Niu & 3\textsuperscript{rd} Yiming Liu \\
\textit{University College London} & \textit{Central South University} & \textit{Nanyang City Fifth Complete School} & \textit{Wuhan Guanggu Future School} \\
\textit{UCL Dept of Physics} & \textit{Dept of Computer Science and Technology} & \textit{High school} & \textit{High school} \\
London, UK & Changsha, China & Nanyang, China & Wuhan, China \\
1176564216@qq.com & lvhuanzhuchat@gmail.com & zyb12668@163.com & lym20070118@outlook.com \\
\end{tabular}
}

\maketitle

\begin{abstract}
 Multi-label sentiment classification plays a vital role in natural language processing by detecting multiple emotions within a single text. However, existing datasets like GoEmotions often suffer from severe class imbalance, which hampers model performance, especially for underrepresented emotions. To address this, we constructed a balanced multi-label sentiment dataset by integrating the original GoEmotions data, emotion-labeled samples from Sentiment140 using a RoBERTa-base-GoEmotions model, and manually annotated texts generated by GPT-4 mini. Our data balancing strategy ensured an even distribution across 28 emotion categories. Based on this dataset, we developed an enhanced multi-label classification model that combines pre-trained FastText embeddings, convolutional layers for local feature extraction, bidirectional LSTM for contextual learning, and an attention mechanism to highlight sentiment-relevant words. A sigmoid-activated output layer enables multi-label prediction, and mixed precision training improves computational efficiency. Experimental results demonstrate significant improvements in accuracy, precision, recall, F1-score, and AUC compared to models trained on imbalanced data, highlighting the effectiveness of our approach. 
\end{abstract}
\maketitle


\section{Introduction}
Sentiment analysis, a core task in natural language processing, systematically identifies and categorizes opinions expressed in text, typically classifying them as positive, negative, or neutral \cite{liu2022sentiment}. With the rapid growth of online communication, it has become indispensable for monitoring public opinion, interpreting customer feedback, and informing strategic decisions. In particular, online reviews can significantly influence public perception, and organizations increasingly rely on automated sentiment analysis to track brand reputation and consumer attitudes \cite{bayhaqy2018sentiment}.

Despite its broad application, sentiment analysis faces persistent challenges. One of the most prominent is data imbalance, where certain emotion categories are severely underrepresented, leading to weaker performance for minority classes \cite{raghunathan2023challenges}. Another difficulty lies in detecting implicit sentiment, such as sarcasm, irony, or wordplay, where the literal meaning of words diverges from the intended message \cite{li2020hemos}. These issues are amplified in multi-label settings, where a single text can express multiple overlapping emotions. Traditional solutions—such as oversampling, undersampling, and cost-sensitive learning—are not universally effective: oversampling can lead to overfitting, undersampling risks losing important context, and cost-sensitive methods require careful parameter tuning to avoid bias.

To address these limitations, this study focuses on optimizing the GoEmotions dataset for balanced, fine-grained multi-label emotion classification. We augment the original dataset with two additional sources: tweets from the Sentiment140 dataset re-labeled into 28 GoEmotions categories using the SamLowe/roberta-base-go-emotions classifier \cite{sam_lowe_2024}, and 20k GPT-4 mini–generated samples designed to enhance long-tail emotion categories. Both sources undergo a multi-step quality control process combining automated verification and manual review. The resulting balanced dataset is then used to train a unified deep learning architecture, enabling controlled comparisons between unbalanced and balanced settings. Experimental results finally show notable improvements in recognizing minority emotions.

\section{Related work}
 Previous studies on sentiment analysis have examined diverse datasets, algorithms, and classification strategies. Zuo \cite{zuo2018sentiment} analyzed 7.7 million Steam game reviews from multiple sources, comparing Gaussian Naïve Bayes and Decision Tree classifiers with TF–IDF features. The Decision Tree achieved 75\% accuracy, benefiting from feature selection through Information Gain or Gini Index. Likewise, Devika et al \cite{devika2016sentiment} compared three sentiment analysis approaches—machine learning, rule-based, and lexical-based—detailing seven machine learning classifiers, and found that machine learning generally yielded higher precision, F1-score, and accuracy.

Bashynska et al. \cite{bashynska2024research} applied a pre-trained BERT model to the GoEmotions dataset, achieving a macro-average F1 score of 0.5070 and demonstrating BERT’s effectiveness in multi-label emotion classification. However, the study did not address dataset imbalance, leading to weaker performance on minority classes. Building on these findings, our work targets improved classification performance while explicitly addressing class imbalance, with the goal of enhancing results for underrepresented emotions and improving applicability in real-world settings.

\section{Methodology}
\subsection{Dataset preparation}\label{3.1}
GoEmotions \cite{demszky2020goemotions} (Kaggle) contains Reddit comments annotated with 28 emotion labels (12 positive, 11 negative, 4 ambiguous, 1 neutral). After removing samples with \texttt{example\_very\_unclear}, 207,814 instances remain. \textcolor{blue}{To mitigate long-tail imbalance, we augmented the pool with (i) Sentiment140 \cite{khasanah2021sentiment} tweets automatically labeled into 28 GoEmotions categories using SamLowe/roberta-base-go-emotions \cite{sam_lowe_2024} (labels stored in the \texttt{model\_labels} field and mapped to the corresponding GoEmotions names; empty sets mapped to \textit{neutral}), and (ii) 20k GPT-4 mini generated texts targeting rare labels. All labels follow the official GoEmotions 28-label taxonomy; no additional semantic remapping was applied beyond string normalization.}

\textcolor{blue}{Quality control combined model-based filtering and human review. Let $p_k(x)$ denote the predicted probability for label $k$ on sample $x$. We keep a sample only if $\max_k p_k(x) \ge 0.7$; labels are assigned by $p_k(x) \ge 0.5$ with a top-3 cap and empty sets discarded. GPT samples follow the same verification before being reviewed by five annotators with majority-vote resolution. The Sentiment140 weak-label pool contains 700,000 tweets, with label cardinality 576,423 (1 label), 115,347 (2 labels), 8,029 (3 labels), and 201 ($\ge 4$ labels). The GPT-4 mini set contains 20,000 texts; the annotation log reports unanimous agreement (5/5) for all 20,000 entries. We audited 200 randomly sampled items per added source, yielding 200/200 matches for Sentiment140 and 200/200 for GPT-4 mini (Table~\ref{tab:audit}). These audits are sample-based and do not preclude residual noise.}

\begingroup
\begin{table}[htbp]
\caption{Audit summary for added sources.}
\label{tab:audit}
\centering
\scriptsize
\begin{tabular}{lccc}
\toprule
\textbf{Source} & \textbf{Pool Size} & \textbf{Audit Size} & \textbf{Match Rate} \\
\midrule
Sentiment140 auto-labels & 700000 & 200 & 1.00 \\
GPT-4 mini (manual labels) & 20000 & 200 & 1.00 \\
\bottomrule
\end{tabular}
\end{table}
\endgroup

\textcolor{blue}{We then merged all sources and applied a greedy balancing procedure with per-label caps: a maximum of 9,000 samples per label, with higher caps for \textit{neutral} (20,000) and \textit{annoyance} (15,000). Multi-label samples update all label counts simultaneously, so exact caps are approximate. The final balanced dataset contains 223,391 instances. Also, we calculate the label cardinality distribution of the new balanced dataset(see \ref{tab:cardinality}), and reports the per-label counts before/after balancing see Table~\ref{tab:label_counts}}
 \begin{figure}[htbp]
\centerline{\includegraphics[width=0.4\textwidth,height=0.2\textheight]{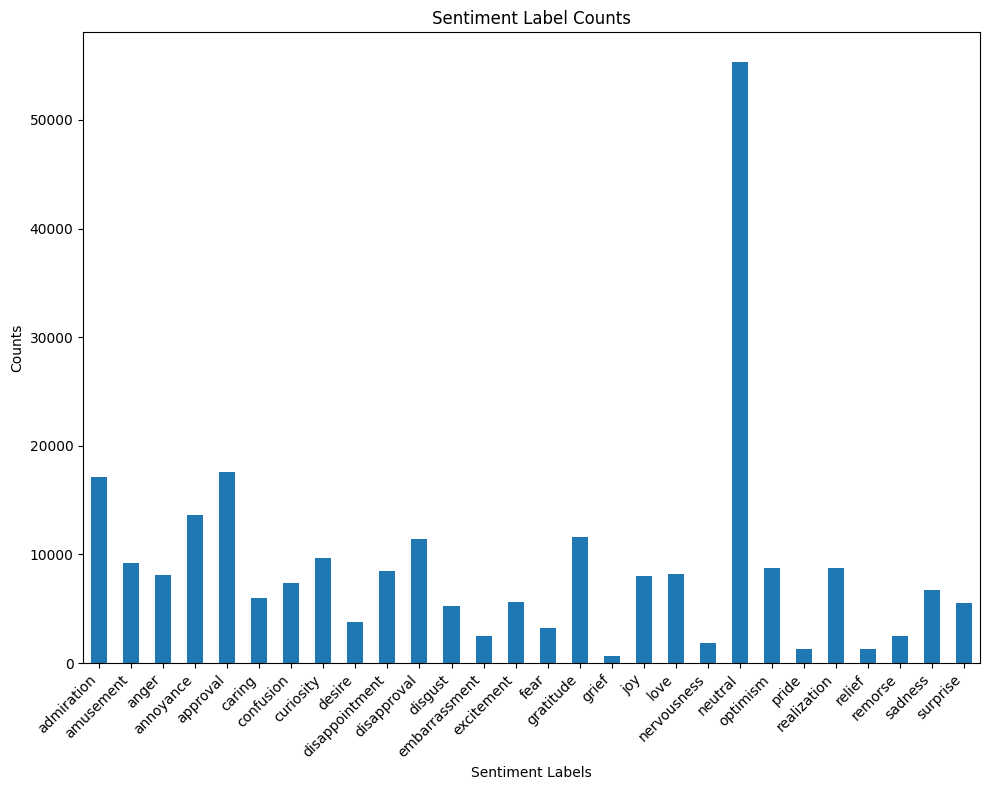}}
\caption{Label distribution of the original GoEmotions dataset after removing \texttt{example\_very\_unclear}.} 
\label{fig:imbalance}
\end{figure}
\begin{figure}[htbp]
\centerline{\includegraphics[width=0.4\textwidth,height=0.2\textheight]{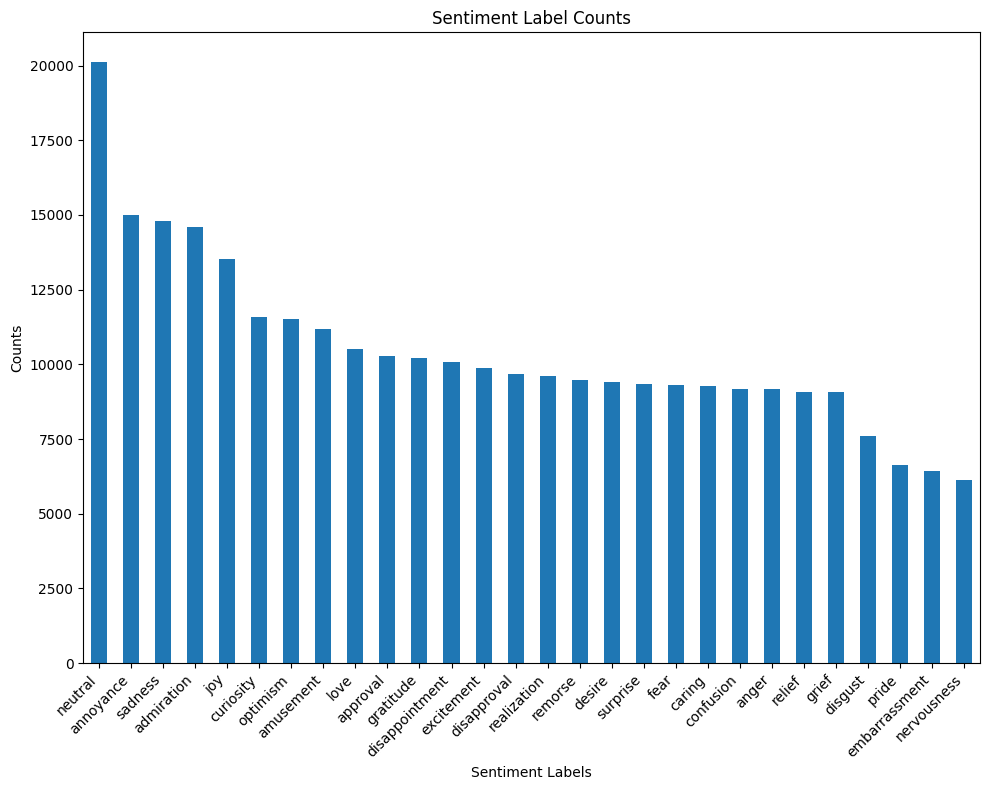}}
\caption{Label distribution after balancing via training-set augmentation (Sentiment140 + GPT-4 mini).}
\label{fig:balanced}
\end{figure}

\begingroup
\color{blue}
\begin{table}[htbp]
\caption{Label cardinality distribution (number of labels per sample).}
\label{tab:cardinality}
\centering
\scriptsize
\begin{tabular}{lrr}
\toprule
\textbf{Labels per Sample} & \textbf{Original} & \textbf{Balanced} \\
\midrule
1 & 171820 (82.68\%) & 160464 (71.83\%) \\
2 & 31187 (15.01\%) & 57299 (25.65\%) \\
3 & 4218 (2.03\%) & 5162 (2.31\%) \\
$\ge$4 & 589 (0.28\%) & 466 (0.21\%) \\
\bottomrule
\end{tabular}
\end{table}
\endgroup

After assembling the dataset, we applied a consistent text preprocessing pipeline. We removed @mentions and URLs, normalized whitespace, and filtered non-textual artifacts. \textcolor{blue}{We then created stratified train/validation/test splits using \textit{MultilabelStratifiedShuffleSplit} (random\_state=42): 178,786 / 22,274 / 22,331 samples (80/10/10). A word-level Tokenizer was fit on the training set only, and all splits were transformed using the same vocabulary with a maximum sequence length of 30 tokens. Emotion labels were converted into 28-dimensional multi-hot vectors using MultiLabelBinarizer. Figure 3 presents the 50 most frequent words in the balanced training set.}

 \begin{figure}[htbp]
\centerline{\includegraphics[width=0.4\textwidth,height=0.2\textheight]{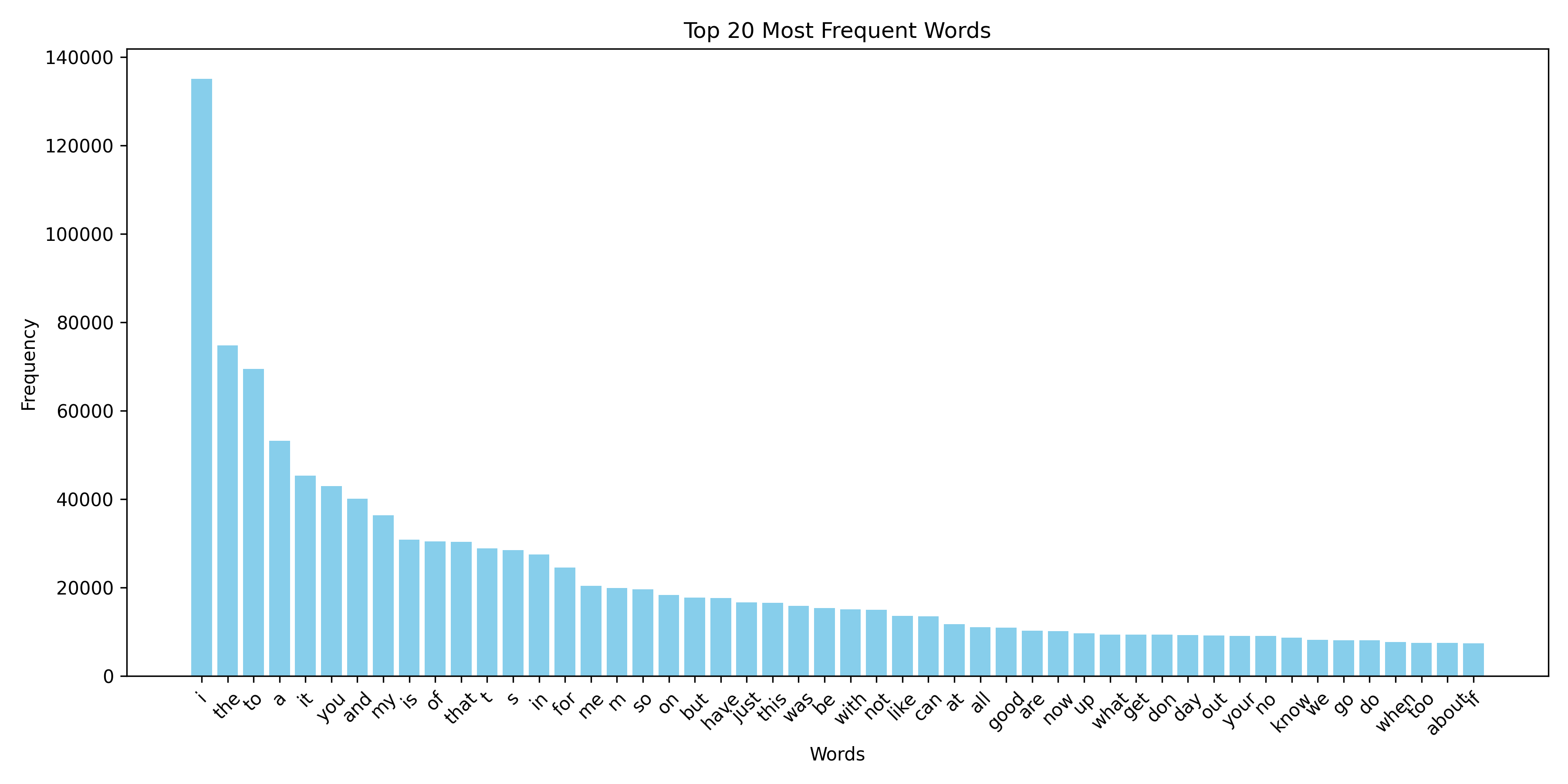}}
\caption{Top-50 most frequent words in the balanced training data.}
\label{fig:count}
\end{figure}

\subsection{Data processing}
\begin{figure}[htbp]
\centerline{\includegraphics[width=0.4\textwidth,height=0.2\textheight]{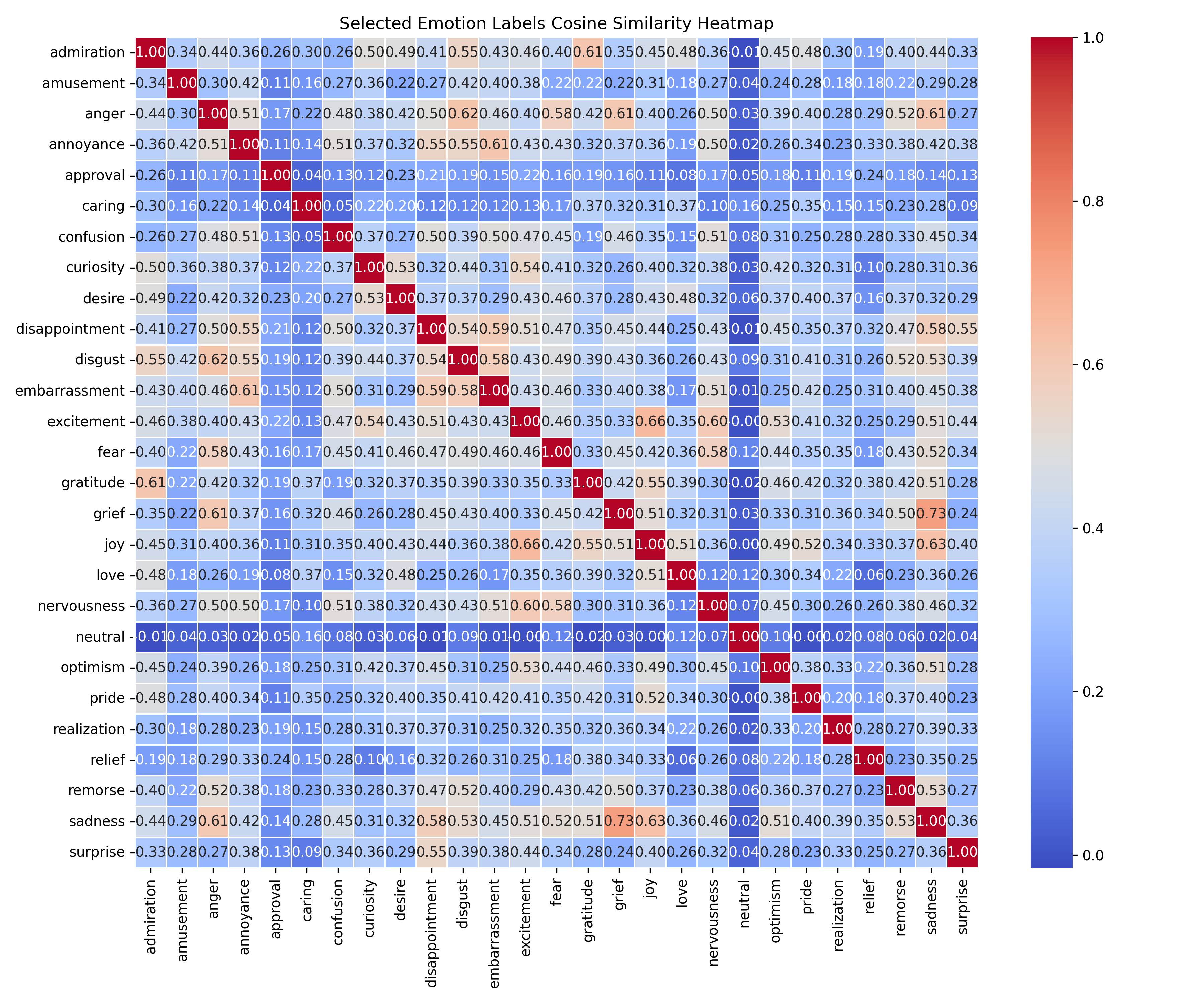}}
\caption{ This figure shows the cosine similarity heat map of emotion labels.  }
\label{fig:heat}
\end{figure}

We compute label co-occurrence statistics on the training split and visualize cosine similarity (Fig.~\ref{fig:heat}). The embedding layer was initialized with a pre-trained FastText matrix to map tokens to 300-dimensional word vectors.
 
After preprocessing, two datasets were fed into the model architecture, which is described in detail in the following section. In brief, the pipeline processes embedded word vectors through convolution, recurrent, and, when applied, attention layers, followed by pooling operations and a sigmoid output layer to produce independent probability scores for each emotion label. The overall workflow is illustrated in Fig.~5.

\begin{figure}[htbp]
    \centering
    \begin{adjustbox}{width=\columnwidth}  
    \begin{tikzpicture}[
        node distance=0.4cm and 0.6cm,
        layer/.style={
            rectangle, 
            draw, 
            fill=blue!10,  
            minimum height=0.6cm, 
            minimum width=2cm,
            text centered,
            font=\bfseries  
        },
        arrow/.style={
            thick,
            ->,
            >=stealth
        }
    ]
        \node[layer] (input) {Collecting Data};
        \node[layer, below=of input] (processing) {Data Processing};
        \node[layer, below=of processing] (features) {Feature Computation};

        \node[layer, below=of features] (embedding_conv) {Embedding Layer + Convolutional Layer (Conv1D)};
        \node[layer, below=of embedding_conv] (bn_pool) {Batch Normalization + Pooling Layer (MaxPooling1D)};
        \node[layer, below=of bn_pool] (lstm_attention) {Bidirectional LSTM + Attention Mechanism};
        \node[layer, below=of lstm_attention] (dense) {Fully Connected Layer (Dense Layer)};

        \node[layer, below=of dense] (metrics) {Accuracy, Precision, Recall, F-score};

        \draw[arrow] (input) -- (processing);
        \draw[arrow] (processing) -- (features);
        \draw[arrow] (features) -- (embedding_conv);
        \draw[arrow] (embedding_conv) -- (bn_pool);
        \draw[arrow] (bn_pool) -- (lstm_attention);
        \draw[arrow] (lstm_attention) -- (dense);
        \draw[arrow] (dense) -- (metrics);

    \end{tikzpicture}
    \end{adjustbox}
    \caption{The step by step workflow of the process of the whole experiment}
\end{figure}

\subsection{Model}
The model starts with a 300-dimensional FastText embedding layer. \textcolor{blue}{We freeze the embedding matrix during training to reduce overfitting and to keep the parameter-efficient setting. Freezing the embedding matrix yields 0.33M trainable parameters out of 21.5M total (Table V). A 1D convolution layer (64 filters, kernel size 5, stride 1) is applied along the token sequence to capture local n-gram patterns, followed by batch normalization and max-pooling to downsample the sequence representation. A bidirectional LSTM (128 units per direction) then models longer-range dependencies. When enabled, an attention module reweights time steps to emphasize emotion-relevant tokens; otherwise, we use temporal average pooling. Finally, a dense layer (128 units) with dropout (0.5) is followed by a sigmoid output layer producing independent probabilities for 28 emotion labels. The full layer-by-layer specification is provided in the Appendix (Table IX).}

\subsection{Training process}
All experiments used the same optimization setup: Adam optimizer (\(\text{lr}=1\times10^{-3}, \beta_{1}=0.9, \beta_{2}=0.999, \epsilon=10^{-7}\)), batch size 256, and 34 training epochs (no early stopping for the ablation curves). Automatic mixed precision (AMP) was enabled to improve training efficiency. For threshold tuning, we performed per-label grid search on the validation set to maximize label-wise F1, using thresholds from 0.05 to 0.95 with a step size of 0.05. The resulting threshold vector \(\tau\) was applied once to the held-out test set for evaluation (no test information was used in tuning).

\textcolor{blue}{To examine the contribution of attention and data balancing, we conducted ablation experiments under three configurations: (i) unbalanced data with attention, (ii) unbalanced data without attention, and (iii) balanced data with attention (final).}

\begin{figure}[htbp]
\centerline{\includegraphics[width=0.5\textwidth,height=0.2\textheight]{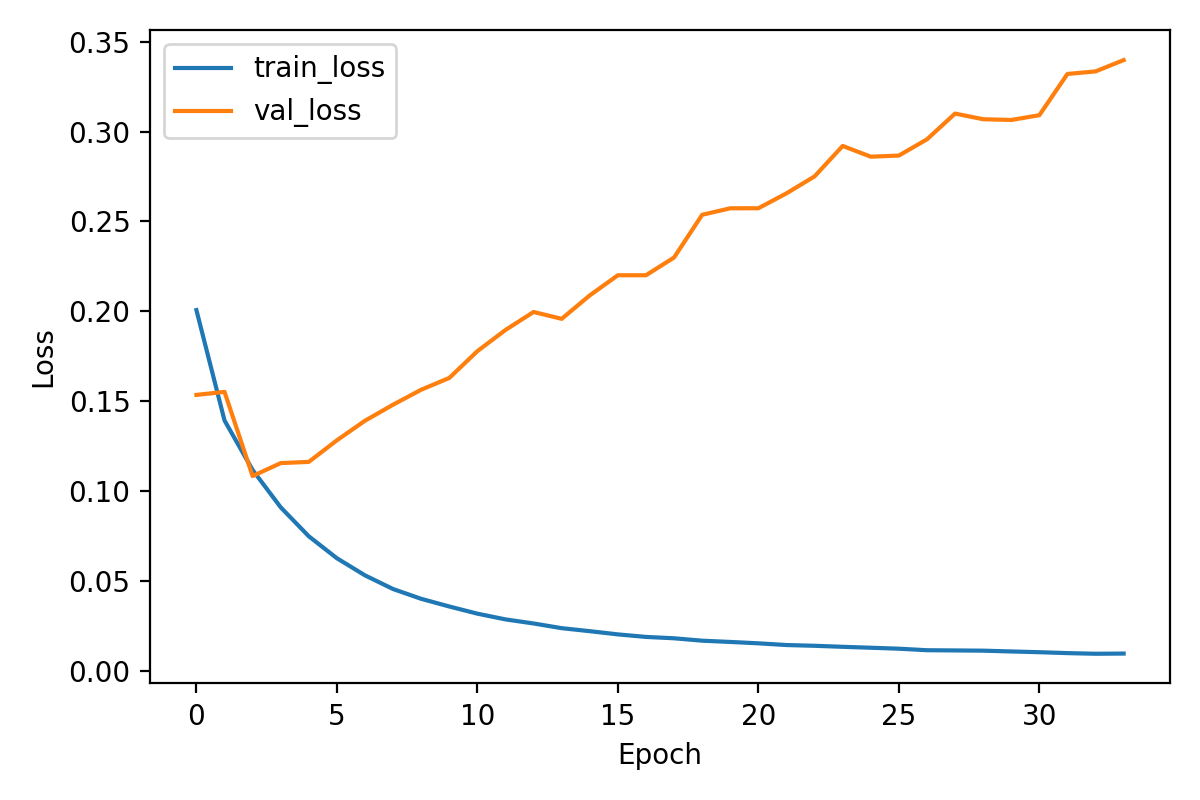}}
\caption{Training and validation loss curves for the unbalanced GoEmotions dataset with attention.}
\label{fig:attn_loss}
\end{figure}

Figure~\ref{fig:attn_loss} shows the unbalanced+attention setting. Training loss decreases steadily, while validation loss starts increasing earlier than training loss, indicating mild overfitting. This divergence is likely driven by the severe label imbalance and weak-label noise in the unbalanced setting, where the model increasingly fits spurious patterns that reduce training loss but do not transfer to validation. In addition, attention can amplify dataset-specific cues, making the model more sensitive to rare-label artifacts and thus accelerating validation loss growth. In this setup, attention reweights token positions to emphasize informative cues for multi-label prediction.

\begin{figure}[htbp]
\centerline{\includegraphics[width=0.5\textwidth,height=0.2\textheight]{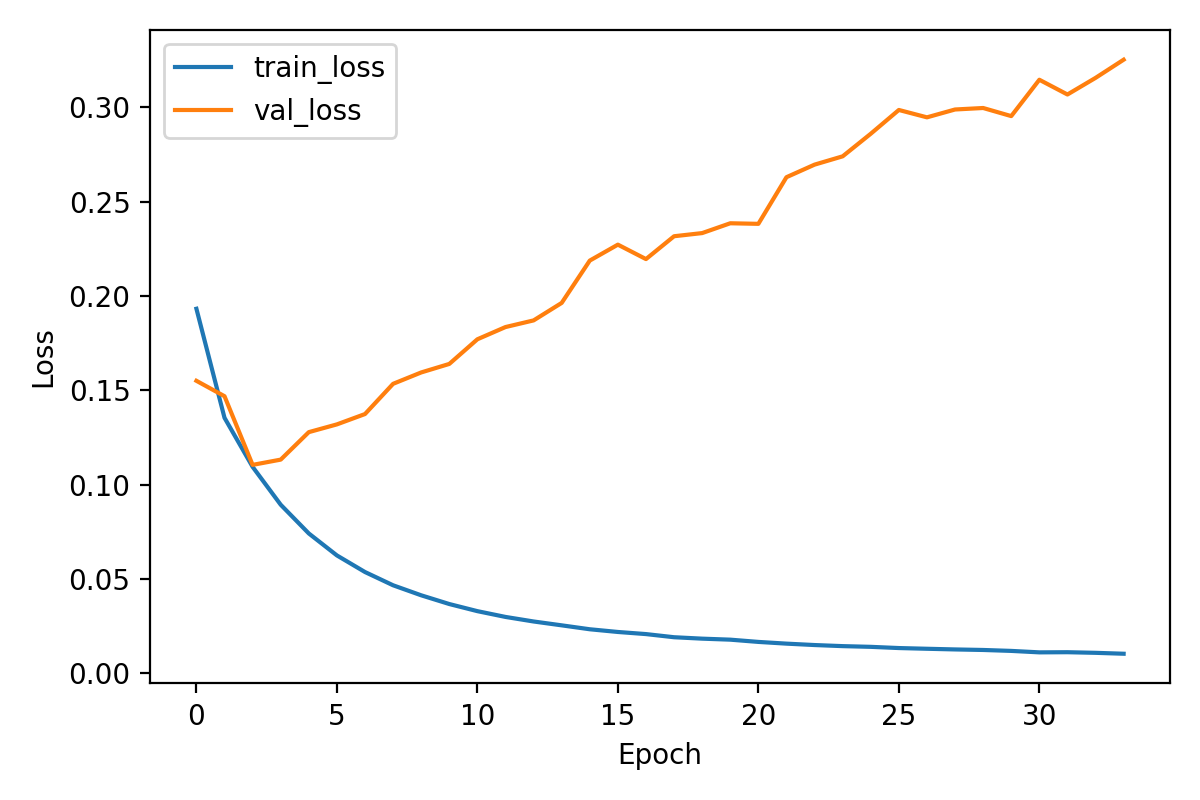}}
\caption{Training and validation loss curves for the unbalanced GoEmotions dataset without attention.}
\label{fig:noattn_loss}
\end{figure}

Figure~\ref{fig:noattn_loss} shows the unbalanced setting without attention. Validation loss reaches its minimum earlier and then increases, indicating stronger overfitting when attention is removed. In this case, temporal average pooling after the BiLSTM is used instead of attention.

\begin{figure}[htbp]
\centerline{\includegraphics[width=0.45\textwidth,height=0.2\textheight]{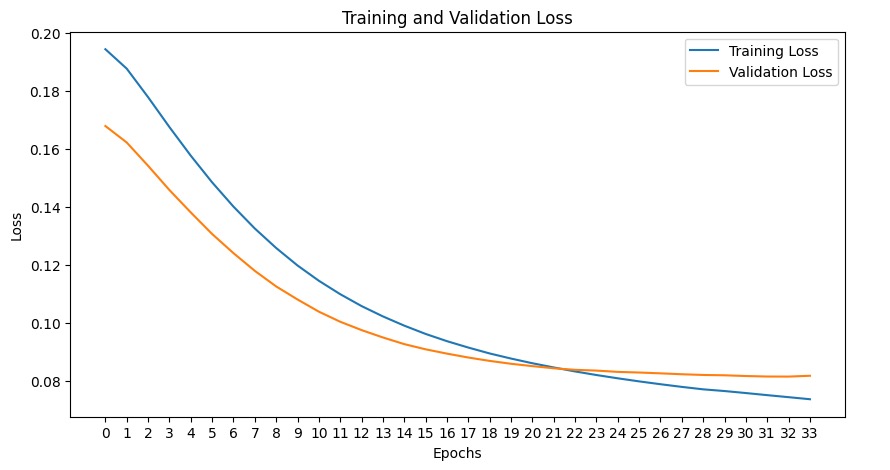}}
\caption{Training and validation loss curves over 34 epochs for the balanced GoEmotions training set with attention (final model).}
\label{fig:baseline_loss}
\end{figure}

Finally, we train the balanced+attention configuration as our final model and use it as the primary reference for subsequent evaluations. Figure~\ref{fig:baseline_loss} reports its training/validation loss curves; both losses decrease and stabilize over epochs, indicating stable optimization under the balanced setting.
\FloatBarrier



\subsection{Evaluation Metrics}
For multi-label classification, we report the following metrics. Unless otherwise stated, predicted labels are obtained by thresholding sigmoid outputs at 0.5.
\textcolor{blue}{\begin{itemize}
    \item \textbf{Subset Accuracy (Exact Match):} the proportion of samples whose predicted label set exactly matches the ground-truth label set.
    \item \textbf{Jaccard Index:} the average, over samples, of the intersection-over-union between predicted and true label sets.
    \item \textbf{Hamming Loss:} the fraction of incorrect label assignments averaged over all sample--label pairs.
    \item \textbf{Micro Precision/Recall/F1:} computed by aggregating true positives, false positives, and false negatives over all labels.
    \item \textbf{Macro Precision/Recall/F1:} the unweighted mean of per-label precision/recall/F1.
    \item \textbf{Macro ROC-AUC:} the mean ROC-AUC computed independently for each label.
\end{itemize}}
\textcolor{blue}{In multi-label settings, the commonly reported accuracy in Table~\ref{tab:model_performance} is the macro average of one-vs-rest label accuracies, which can be high because most labels are negative for most samples. For the balanced+attention model, macro per-label accuracy is \(\sim 0.966\) at a 0.5 threshold and \(\sim 0.946\) after threshold optimization, while subset accuracy remains around 0.31. We therefore emphasize Jaccard and F1 as primary indicators.}
\subsection{Implementation}
We implemented all experiments in TensorFlow 2.x (Keras), using NumPy and Pandas for data handling, scikit-learn for evaluation utilities (e.g., MultiLabelBinarizer and metric computation), Gensim to load pre-trained FastText embeddings, and NLTK for text preprocessing. Experiments were run on an NVIDIA Tesla V100 GPU (16\,GB) with an Intel Xeon CPU and 64\,GB RAM. For reproducibility, we fix the stratified split with \texttt{random\_state=42}; other stochastic components follow framework defaults.

\section{Analysis and Discussion}
\subsection{Baseline vs Ablation study}
To evaluate multi-label classification, we report subset (exact-match) accuracy, Jaccard index, micro-/macro-averaged F1, and macro-averaged ROC-AUC under a fixed decision threshold of 0.5 for consistent comparison across ablations. \textcolor{blue}{ We use subset accuracy (rather than label-wise accuracy) because the latter can be inflated by the large number of negative labels in multi-label settings. Table I summarizes the ablation results under this fixed-threshold protocol.}
For the balanced setting in Table~\ref{tab:ablation}, we use an oversampled variant of the GoEmotions training split where positive labels are oversampled to the maximum per-label positive count to isolate the effect of balancing.
\begingroup
\color{blue}
\begin{table}[htbp]
\centering
\small
\caption{Ablation study results on the GoEmotions test set (threshold = 0.5).}
\resizebox{\columnwidth}{!}{
\begin{tabular}{lccccc}
\toprule
\textbf{Setting} & \textbf{Subset Acc} & \textbf{Jaccard} & \textbf{Micro-F1} & \textbf{Macro-F1} & \textbf{Macro AUC} \\
\midrule
Unbalanced + Attn      & 0.305 & 0.334 & 0.431 & 0.164 & 0.821 \\
Unbalanced - Attn      & 0.237 & 0.263 & 0.372 & 0.149 & 0.820 \\
Balanced + Attn     & 0.311 & 0.341 & 0.438 & 0.201 & 0.829 \\
\bottomrule
\end{tabular}
}
\label{tab:ablation}
\end{table}
\endgroup

\textcolor{blue}{Table~\ref{tab:ablation} shows that attention improves performance on the unbalanced data, and balancing further boosts all metrics (e.g., Macro-F1 from 0.164 to 0.201). Our final model uses the balanced+attention setting. We also report per-label F1 (threshold=0.5) for a fine-grained comparison across emotions, see following figures.}
\begin{figure}[!htbp]
  \centering
  \includegraphics[width=0.45\textwidth]{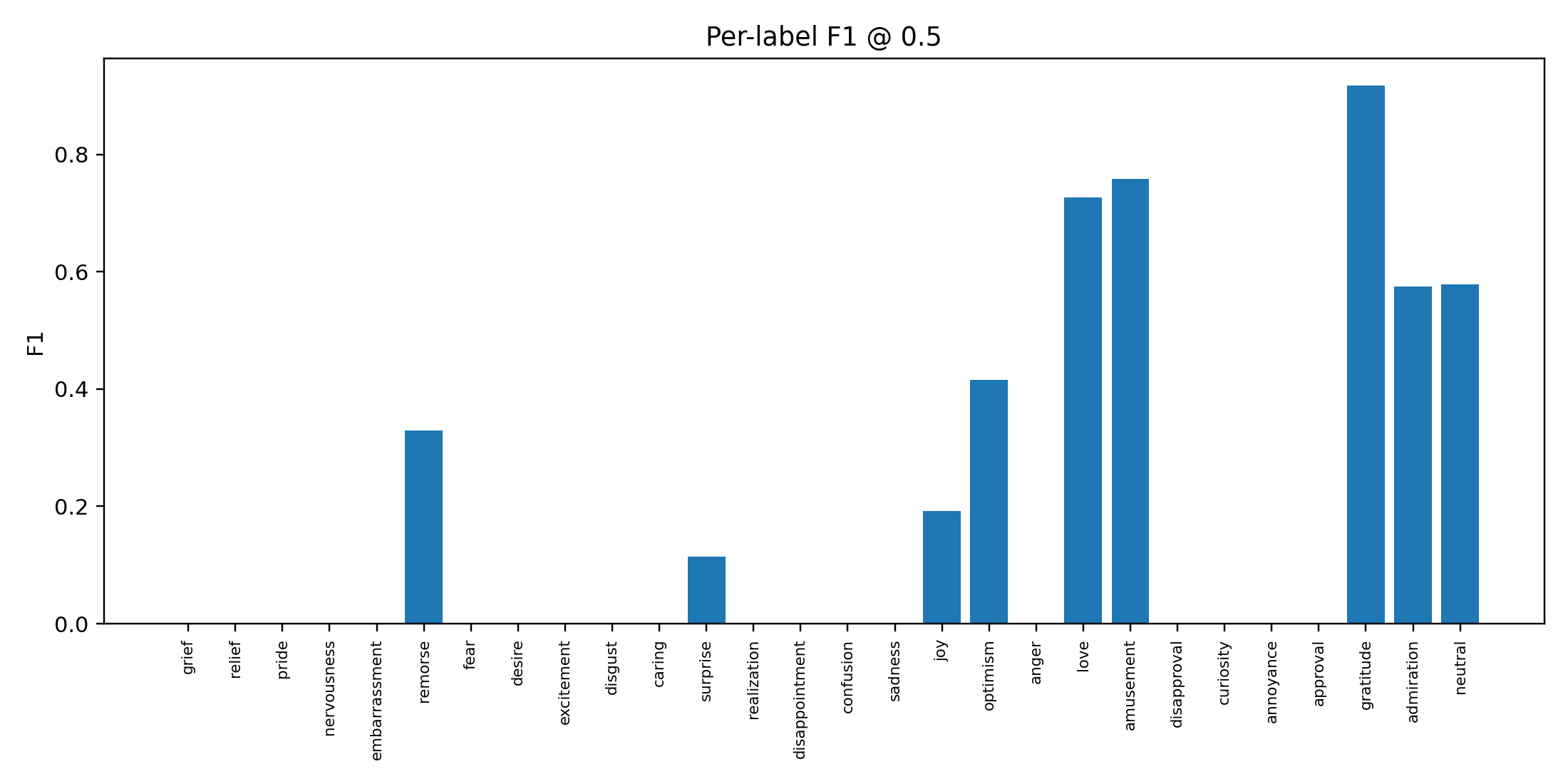}
  \caption{Per-label F1-scores (0.5 threshold) for the Unbalanced + Attention configuration.}
  \label{fig:f1_unbalanced_attn}
\end{figure}
\noindent
\begin{figure}[!htbp]
  \centering
  \includegraphics[width=0.45\textwidth]{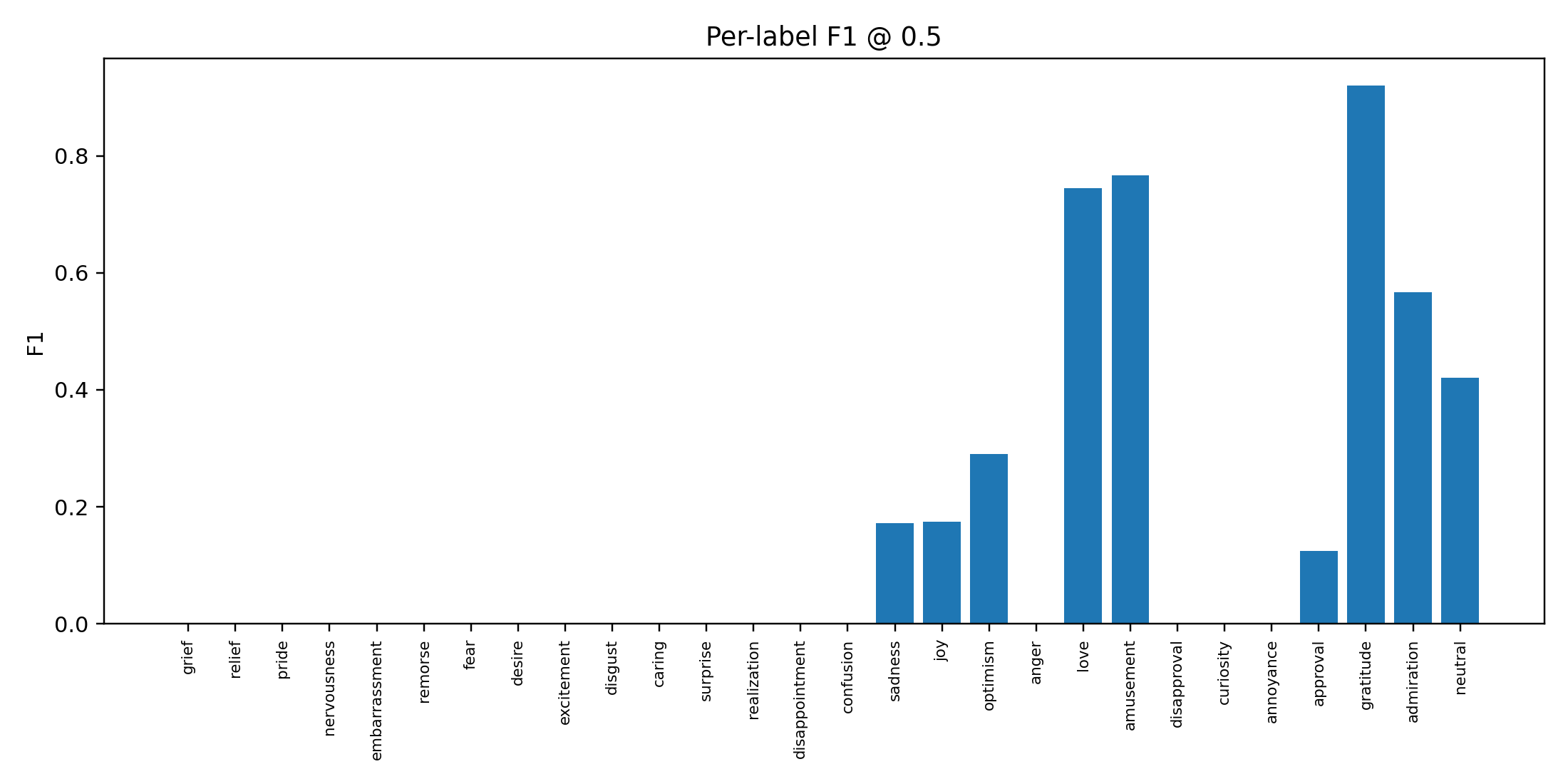}
  \caption{Per-label F1-scores (0.5 threshold) for the Unbalanced - Attention configuration.}
  \label{fig:f1_unbalanced_noattn}
\end{figure}

\begin{figure}[!htbp]
  \centering
  \includegraphics[width=0.45\textwidth]{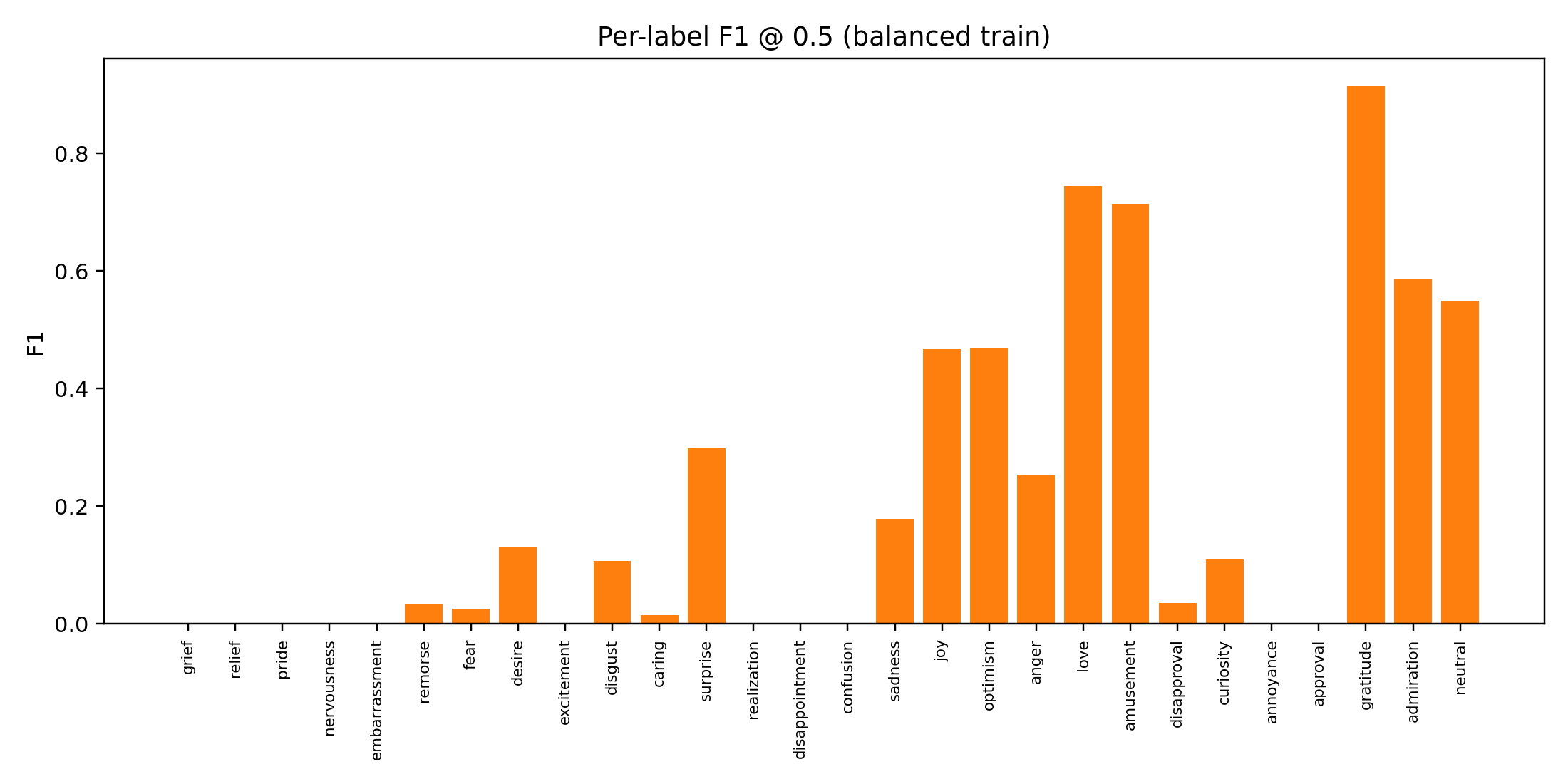}
  \caption{Per-label F1-scores (0.5 threshold) for the Balanced + Attention configuration.}
  \label{fig:f1_balanced_attn}
\end{figure}
\FloatBarrier
We next compare our final balanced+attention model against transformer baselines.

\subsection{Comparison with Transformer Baselines}
\begin{figure}[htbp]
  \centering
  \includegraphics[width=0.45\textwidth]{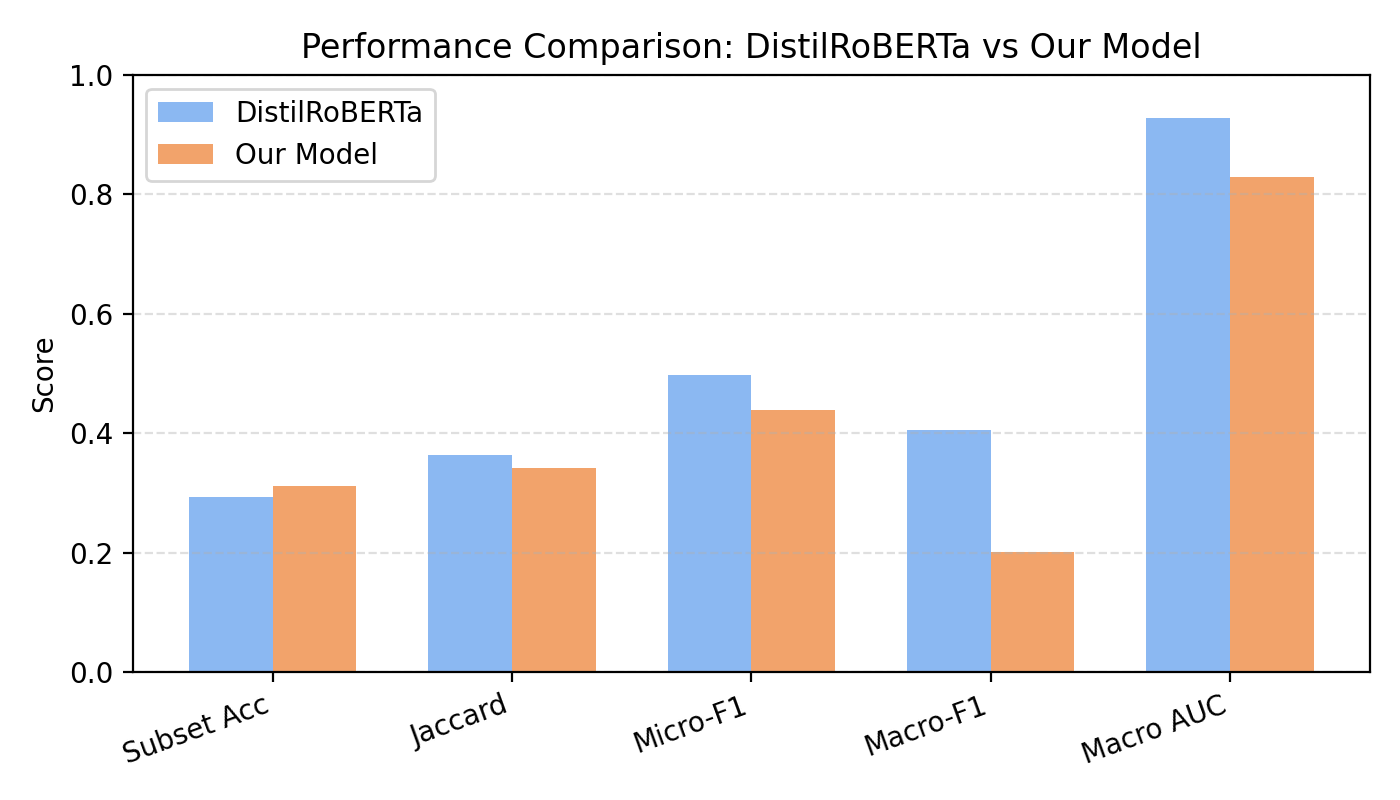}
  \caption{Bar chart comparison of DistilRoBERTa and our model (threshold = 0.5).}
  \label{fig:transformer_bar}
\end{figure}

\textcolor{blue}{We fine-tuned a lightweight DistilRoBERTa model on the balanced dataset using the same stratified split (train/val/test = 178,786/22,274/22,331). Table~\ref{tab:transformer_comparison} reports results at threshold 0.5, Table~\ref{tab:transformer_gap} summarizes the gap relative to our model, and Fig.~\ref{fig:transformer_bar} visualizes the comparison.}

\begingroup
\color{blue}
\begin{table}[htbp]
\centering
\caption{DistilRoBERTa baseline on the balanced dataset (threshold = 0.5).}
\resizebox{\columnwidth}{!}{
\begin{tabular}{lccccc}
\toprule
\textbf{Model} & \textbf{Subset Acc} & \textbf{Jaccard} & \textbf{Micro-F1} & \textbf{Macro-F1} & \textbf{Macro AUC} \\
\midrule
DistilRoBERTa (balanced) & 0.293 & 0.363 & 0.498 & 0.405 & 0.928 \\
\bottomrule
\end{tabular}}
\label{tab:transformer_comparison}
\end{table}
\endgroup

\begingroup
\color{blue}
\begin{table}[htbp]
\centering
\caption{Gap analysis between DistilRoBERTa and our model (threshold = 0.5).}
\resizebox{\columnwidth}{!}{
\begin{tabular}{lccc}
\toprule
\textbf{Metric} & \textbf{Our Model} & \textbf{DistilRoBERTa} & \textbf{Delta} \\
\midrule
Subset Accuracy & 0.311 & 0.293 & -0.018 \\
Jaccard & 0.341 & 0.363 & +0.022 \\
Micro-F1 & 0.438 & 0.498 & +0.060 \\
Macro-F1 & 0.201 & 0.405 & +0.204 \\
Macro AUC & 0.829 & 0.928 & +0.099 \\
\bottomrule
\end{tabular}}
\label{tab:transformer_gap}
\end{table}
\endgroup

\textcolor{blue}{DistilRoBERTa improves Micro-F1, Macro-F1, and Macro-AUC relative to our lightweight model (Table~\ref{tab:transformer_gap}), while subset accuracy remains comparable. The largest gap is in Macro-F1 (+0.204), which is expected given the stronger representation capacity of Transformers. Our contribution is the data balancing pipeline and an efficient architecture that improves performance relative to unbalanced training (Table~\ref{tab:ablation}). Under per-label thresholds, DistilRoBERTa reaches Micro-F1 0.645 and Macro-F1 0.647.}

\subsection{Single-emotion performance evaluation}
As an auxiliary analysis, we performed per-label threshold optimization: for each label, we selected a decision threshold on the validation set to maximize label-wise F1, producing a threshold vector \(\tau\). This vector was applied once to the held-out test set for evaluation (no test information was used in tuning). Since this corresponds to a different operating point, these results are reported separately in Table~\ref{tab:model_performance}.\textcolor{blue}{(see Appendix)}

Table~\ref{tab:model_performance} summarizes per-label performance under the optimized thresholds. Note that per-label accuracy is computed in a one-vs-rest manner and can be inflated by true negatives; we therefore emphasize F1 and AUC for interpretation. Overall, the model achieves consistently high ROC-AUC across labels, while performance varies in F1 for semantically subtle and context-dependent emotions (e.g., approval and disappointment), which often overlap with related categories and exhibit higher annotation ambiguity.

To illustrate model behavior, Table~\ref{tab:examples} lists the top predicted labels and probabilities for representative sentences. The model assigns a dominant label when the emotion is explicit (e.g., sadness in ``I feel bad for those boys'') and can capture co-occurring emotions in multi-faceted sentences (e.g., ``good news ... bad news ...''). More colloquial or implicit expressions lead to more dispersed probabilities across labels, highlighting the challenge of ambiguity and context in multi-label emotion recognition.

\subsection{Evaluation and Discussion}
Overall, our experiments show that alleviating label imbalance improves multi-label emotion classification relative to training on the original unbalanced data. By augmenting the GoEmotions training split with additional samples from Sentiment140 and GPT-4 mini, the balanced setting yields consistent gains in Jaccard and F1 metrics compared with the unbalanced counterparts (Table~\ref{tab:ablation}), indicating improved coverage of long-tail emotions. Moreover, adding the attention module further improves performance over the no-attention variant, suggesting that reweighting informative token positions helps capture salient emotional cues in short texts.
Error analysis indicates that rare or ambiguous labels remain the most challenging. In the balanced+attention setting, labels such as \textit{grief}, \textit{relief}, \textit{nervousness}, and \textit{embarrassment} exhibit the lowest per-label F1 scores under optimized thresholds, while frequent labels such as \textit{gratitude}, \textit{love}, and \textit{neutral} achieve substantially higher F1. This pattern suggests that even after balancing, semantic ambiguity and weak supervision still affect minority labels.

Despite these gains, several challenges remain. Some emotions (e.g., approval and disappointment) are semantically subtle and highly context-dependent, often overlapping with related categories and thus yielding lower per-label F1 (Table~\ref{tab:model_performance}). In addition, weak supervision (Sentiment140 re-labeling) and generated texts can introduce annotation noise, which may limit generalization. Our noise estimates are based on 200-sample audits per added source; larger audits could further quantify residual noise. This motivates stronger auditing and noise-aware training strategies, such as stricter filtering, double-annotation of sampled subsets, and robustness-oriented losses. Finally, although our CNN--BiLSTM--attention model is lighter than Transformer-based approaches, training with augmented data still incurs non-trivial computational cost.

Future work will focus on (i) stronger Transformer variants trained with the same balancing protocol, (ii) expanded data quality audits and calibration for per-label thresholds, and (iii) extending the balancing pipeline to multilingual and other multi-label tasks (e.g., topic or intent detection), where long-tail labels are also common.

\section{Conclusion}
This paper investigates multi-label emotion classification under severe class imbalance. We propose a data balancing pipeline that augments the GoEmotions training split with additional weakly labeled and generated samples, and we evaluate a lightweight CNN--BiLSTM model with an attention module. Experiments show that balancing improves multi-label metrics relative to training on the original unbalanced data, and attention further contributes to performance, particularly for underrepresented emotions.

Beyond performance gains, we provide a reproducible training and evaluation protocol, including validation-only threshold tuning to avoid test leakage. The main limitations are the potential noise introduced by weak supervision and generated data, and the computational cost of training on enlarged datasets.

Future work will focus on stronger quality auditing and noise-robust learning, larger Transformer variants under the same split/tuning protocol, and extending the proposed balancing strategy to multilingual and other multi-label text classification tasks (e.g., topic or intent detection).

\section{Acknowledgement}
We express our sincere gratitude to Professor Björn Schuller for his valuable feedback, patience, and continuous support throughout this research. Special thanks are also extended to teaching assistant Weihao Tang for his insightful guidance during project development.
Zijin Su and Huanzhu Lv contributed equally to this work and are recognized as co-first authors. 

\section{Author Contributions}
Zijin Su and Huanzhu Lv contributed equally to this work and are recognized as co-first authors. Zijin Su led the conceptualization, data curation, formal analysis, investigation, and project administration, and contributed to both the original draft and its revision. Huanzhu Lv supported the conceptualization, methodology, resources, and validation, and contributed to data analysis and manuscript editing. Yuren Niu contributed to visualization and overall review of the manuscript. Yiming Liu provided resources and participated in the revision of the manuscript. All authors have read and approved the final version of the manuscript.

\section{Funding}
This research received no external funding. 
\section{Conflict of interest}
The authors declare that they have no conflict of interest.
\section{Data Availability}
The balanced dataset (\texttt{final\_balanced\_df\_output.csv} / \texttt{balanced\_emotion\_dataset.csv}), the Sentiment140 auto-labeled pool (\texttt{sentiment140\_with\_go\_emotions\_labels.csv}), audit logs (\texttt{sentiment140\_noise\_audit.csv}, \texttt{gpt4mini\_annotations.csv}, \texttt{gpt4mini\_audit.csv}), label count tables, and ablation summaries that support the findings of this study are provided with this revision. The core dataset and figures are also openly available in Zenodo at https://doi.org/10.5281/zenodo.15837871 and https://doi.org/10.5281/zenodo.16890154, https://doi.org/10.5281/zenodo.18366640(records 15837871, 16890154 and 18366640),.

\bibliography{ref}
\section{Appendix}
For reproducibility, we provide the full per-label threshold optimization results, raw evaluation outputs, and additional plots at the following repository: https://zenodo.org/records/16890154. Table III and the detailed model specification are shown on the next page.
\begingroup
\color{blue}
\begin{table*}[htbp]
\caption{Label counts before and after balancing.}
\label{tab:label_counts}
\centering
\scriptsize
\begin{tabular}{lrrlrr}
\toprule
\textbf{Emotion} & \textbf{Original} & \textbf{Balanced} & \textbf{Emotion} & \textbf{Original} & \textbf{Balanced} \\
\midrule
admiration & 17131 & 14603 & amusement & 9245 & 11188 \\
anger & 8084 & 9176 & annoyance & 13618 & 15001 \\
approval & 17620 & 10285 & caring & 5999 & 9290 \\
confusion & 7359 & 9185 & curiosity & 9692 & 11580 \\
desire & 3817 & 9404 & disappointment & 8469 & 10085 \\
disapproval & 11424 & 9690 & disgust & 5301 & 7586 \\
embarrassment & 2476 & 6422 & excitement & 5629 & 9868 \\
fear & 3197 & 9314 & gratitude & 11625 & 10225 \\
grief & 673 & 9065 & joy & 7983 & 13518 \\
love & 8191 & 10510 & nervousness & 1810 & 6129 \\
optimism & 8715 & 11515 & pride & 1302 & 6621 \\
realization & 8785 & 9607 & relief & 1289 & 9087 \\
remorse & 2525 & 9476 & sadness & 6758 & 14789 \\
surprise & 5514 & 9337 & neutral & 55298 & 20107 \\
\bottomrule
\end{tabular}
\end{table*}
\endgroup

\begin{table*}[!htbp]
\centering  
\caption{\textcolor{blue}{Per-label performance under validation-tuned thresholds (accuracy is one-vs-rest per label).}}
\label{tab:model_performance}
\resizebox{\textwidth}{!}{  
\begin{tabular}{lccccccc}  
\toprule
\textbf{Emotion} & \textbf{Accuracy} & \textbf{Precision} & \textbf{Recall} & \textbf{F1-Score} & \textbf{AUC} & \textbf{Support} & \textbf{Threshold} \\ 
\midrule
Admiration   & 0.9596 & 0.7079 & 0.6628 & 0.6846 & 0.9521 & 1460 & 0.35 \\
Amusement    & 0.9843 & 0.8389 & 0.8480 & 0.8435 & 0.9814 & 1119 & 0.25 \\
Anger        & 0.9696 & 0.8398 & 0.5585 & 0.5964 & 0.9454 & 918  & 0.35 \\
Annoyance    & 0.9376 & 0.5327 & 0.6216 & 0.5737 & 0.9343 & 1500 & 0.35 \\
Approval     & 0.9507 & 0.4855 & 0.4511 & 0.4216 & 0.8969 & 1029 & 0.20 \\
Caring       & 0.9723 & 0.7011 & 0.6549 & 0.6771 & 0.9575 & 929  & 0.25 \\
Confusion    & 0.9731 & 0.7656 & 0.5476 & 0.6385 & 0.9494 & 919  & 0.35 \\
Curiosity    & 0.9555 & 0.5866 & 0.5580 & 0.5719 & 0.9137 & 1158 & 0.20 \\
Desire       & 0.9824 & 0.8111 & 0.7423 & 0.7752 & 0.9723 & 940  & 0.40 \\
Disappointment & 0.9553 & 0.5411 & 0.4660 & 0.5008 & 0.9116 & 1009 & 0.20 \\
Disapproval  & 0.9541 & 0.4786 & 0.5496 & 0.5117 & 0.9328 & 969  & 0.35 \\
Disgust      & 0.9646 & 0.4933 & 0.5674 & 0.5278 & 0.9413 & 759  & 0.30 \\
Embarrassment & 0.9851 & 0.8581 & 0.5814 & 0.6932 & 0.9422 & 642  & 0.40 \\
Excitement   & 0.9692 & 0.6580 & 0.6361 & 0.6468 & 0.9502 & 987  & 0.25 \\
Fear         & 0.9842 & 0.8086 & 0.7978 & 0.8031 & 0.9751 & 931  & 0.55 \\
Gratitude    & 0.9904 & 0.9074 & 0.8737 & 0.8903 & 0.9841 & 1023 & 0.40 \\
Grief        & 0.9918 & 0.9336 & 0.8480 & 0.8888 & 0.9909 & 896  & 0.25 \\
Joy          & 0.9691 & 0.7698 & 0.6833 & 0.7239 & 0.9645 & 1352 & 0.30 \\
Love         & 0.9858 & 0.8350 & 0.8649 & 0.8497 & 0.9796 & 1051 & 0.35 \\
Nervousness  & 0.9859 & 0.8548 & 0.5906 & 0.6986 & 0.9466 & 613  & 0.40 \\
Neutral      & 0.9155 & 0.5135 & 0.5466 & 0.5295 & 0.9045 & 2011 & 0.30 \\
Optimism     & 0.9776 & 0.8361 & 0.7190 & 0.7731 & 0.9642 & 1152 & 0.30 \\
Pride        & 0.9942 & 0.9924 & 0.8065 & 0.8898 & 0.9667 & 662  & 0.20 \\
Realization  & 0.9647 & 0.5734 & 0.4711 & 0.5172 & 0.9143 & 961  & 0.20 \\
\bottomrule
\end{tabular}
}
\end{table*}

\begin{table*}[htbp]
\centering
\caption{Examples of top predicted emotion labels and probabilities (sigmoid outputs).}
\label{tab:examples}
\resizebox{\textwidth}{!}{%
\begin{tabular}{p{7cm}cccc}
\toprule
\textbf{Sentence} & \textbf{Top-1} & \textbf{Top-2} & \textbf{Top-3} & \textbf{Top-4} \\
\midrule
wants to fall in love & desire: 0.98 & love: 0.86 & - & - \\
My mom liked her mothers day gift SUCCESS & love: 0.82 & excitement: 0.63 & - & - \\
yes hand of fatima or hamsa i am tryna start a collection if you find one great if you don t it s cool just have FUN & joy: 0.72 & optimism: 0.55 & amusement: 0.42 & - \\
I feel bad for those boys & sadness: 0.93 & - & - & - \\
People done be ignorant & neutral: 0.54 & disappointment: 0.34 & - & - \\
good news TONS of new people signing up for my virtual cocktail party thursday bad news my site just went down & surprise: 0.73 & sadness: 0.68 & joy: 0.65 & disappointment: 0.55 \\
\bottomrule
\end{tabular}}
\end{table*}

\begin{table*}[htbp]  
\centering
\caption{Model: Text\_Sentiment\_Classification\_Model}
\begin{tabular*}{\textwidth}{@{\extracolsep{\fill}} lcccc}  
\toprule
\textbf{Layer (type)} & \textbf{Output Shape} & \textbf{Param \#} & \textbf{Connected to} \\
\midrule
\textcolor{blue}{input\_layer (InputLayer)} & (None, 30) & 0 & - \\
\textcolor{blue}{embedding\_layer (Embedding)} & (None, 30, 300) & 21,210,600 & input\_layer[0][0] \\
\textcolor{blue}{conv\_layer (Conv1D)} & (None, 26, 64) & 96,064 & embedding\_layer[0][0] \\
\textcolor{blue}{batch\_normalization (BatchNormalization)} & (None, 26, 64) & 256 & conv\_layer[0][0] \\
\textcolor{blue}{max\_pooling1d (MaxPooling1D)} & (None, 13, 64) & 0 & batch\_normalization[0][0] \\
\textcolor{blue}{bidirectional (Bidirectional)} & (None, 13, 256) & 197,632 & max\_pooling1d[0][0] \\
\textcolor{blue}{dense (Dense)} & (None, 13, 1) & 257 & bidirectional[0][0] \\
\textcolor{blue}{lambda (Lambda)} & (None, 13) & 0 & dense[0][0] \\
\textcolor{blue}{activation (Activation)} & (None, 13) & 0 & lambda[0][0] \\
\textcolor{blue}{lambda\_1 (Lambda)} & (None, 13, 1) & 0 & activation[0][0] \\
\textcolor{blue}{multiply (Multiply)} & (None, 13, 256) & 0 & bidirectional[0][0], lambda\_1[0][0] \\
\textcolor{blue}{lambda\_2 (Lambda)} & (None, 256) & 0 & multiply[0][0] \\
\textcolor{blue}{dense\_layer (Dense)} & (None, 128) & 32,896 & lambda\_2[0][0] \\
\textcolor{blue}{dropout (Dropout)} & (None, 128) & 0 & dense\_layer[0][0] \\
\textcolor{blue}{output\_layer (Dense)} & (None, 28) & 3,612 & dropout[0][0] \\
\midrule
\textbf{Total params} & 21,541,317 & (82.17 MB) & \\
\textbf{Trainable params} & 330,589 & (1.26 MB) & \\
\textbf{Non-trainable params} & 21,210,728 & (80.91 MB) & \\
\bottomrule
\end{tabular*}
\end{table*}
\end{document}